\documentclass{article}

\usepackage[dblblindworkshop,final]{neurips_2026}

\usepackage[utf8]{inputenc}
\usepackage[T1]{fontenc}
\usepackage{courier}
\usepackage{url}
\usepackage{booktabs}
\usepackage{graphicx}
\usepackage{float}
\usepackage{pdflscape}
\usepackage{amsmath}
\usepackage{amssymb}
\usepackage{nicefrac}
\usepackage{microtype}
\usepackage{xcolor}
\usepackage{hyperref}
\hypersetup{hidelinks}

\workshoptitle{Evaluation of Interactive Agents}
\title{DetourBench: Trajectory-Level Evaluation of Interactive Agents
with Exact Minimum-Action Bounds}

\hypersetup{pdftitle={DetourBench: Trajectory-Level Evaluation of Interactive Agents with Exact Minimum-Action Bounds},pdfauthor={Hanlin Tian, Sihan Zhu, Zhao Yang, Yuxiang Wang, HU ZHIQUAN, Yu Mi, Hongquan Zhu, Qiufei Hu}}

\author{%
  Hanlin Tian$^{1}$ \quad Sihan Zhu$^{2}$ \quad Zhao Yang$^{3}$ \quad Yuxiang Wang$^{4}$ \\
  \textbf{Zhiquan Hu$^{5}$ \quad Yu Mi$^{6}$ \quad Hongquan Zhu$^{1}$ \quad Qiufei Hu$^{1}$} \\[3pt]
  \small $^{1}$Datagrids.ai \quad $^{2}$Northwest Polytechnical University Xi'an \\
  \small $^{3}$Nanyang Technological University \quad $^{4}$East China Normal University \\
  \small $^{5}$Tongji University \quad $^{6}$Peking University \\
  \small \texttt{tianhl@datagrids.ai}
}

\begin{document}

\maketitle

\begin{abstract}
Benchmarks judge language-model agents mainly by whether they reach the correct
final state, not by how they got there. Endpoint scoring cannot separate an
agent that retrieved exactly the evidence a decision required from one that read
most of the environment, nor an agent that never located the evidence from one
that stopped a single action short. Settling this needs a reference cost, and
authored solutions establish only that a task is solvable, not that no shorter
route exists.

We present DetourBench, which constrains the environment until that minimum is
computable. Each task is a finite, partially observable transition system with
unit-cost accesses whose resources become globally addressable after discovery.
Exhaustive search yields the full-information minimum action count $L_o$ with a
replayable witness; from graph-recoverable terminal states it also yields exact
residual work $R_T$, which separates an episode that stopped one access short
from one that wandered, while an irreversible wrong commitment is recorded as
having no finite residual distance rather than assigned one. Success,
attainment, residual progress, and normalized detour are reported separately.
Across 375 scored episodes --- five model endpoints, 25 integrated tasks, three
repeats --- success ranges from 58.7\% to 96.0\% while conditional attainment
ranges from 0.803 to 0.890 and orders the models differently. Among
recoverable failures, residual work separates near-complete trajectories from
costly incompletion and from irreversible commitment. We open-source the instances, oracle witnesses, and
evaluation harness at \href{https://github.com/DatagridsAI/DetourBench}{%
\textcolor{blue}{https://github.com/DatagridsAI/DetourBench}}.
\end{abstract}

\section{Introduction}

\begin{figure}[t]
\centering
\includegraphics[width=0.58\textwidth]{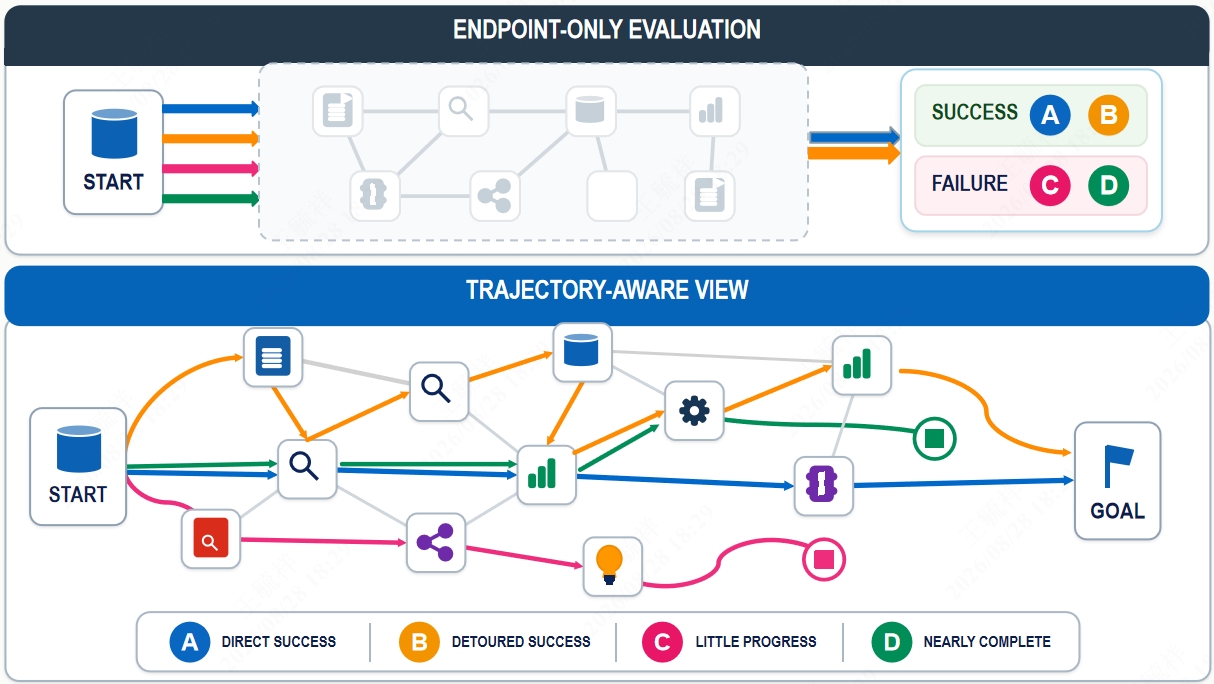}
\caption{\textbf{Endpoint evaluation versus trajectory-aware measurement.}
Endpoint success conflates direct and circuitous successful trajectories and
assigns the same outcome to recoverable near-misses and irreversible failures.
DetourBench compares charged accesses with the full-information minimum $L_o$
and computes residual work $R_T$ from recoverable terminal states, separating
outcome, oracle-relative excess interaction, and residual-distance progress.}
\label{fig:teaser}
\end{figure}

Interactive-agent benchmarks commonly score whether an agent completed the
task \citep{zhou2023webarena,xie2024osworld,mialon2023gaia,
trivedi2024appworld,swebenchverified}. While necessary, endpoint scoring is
silent about process (Figure~\ref{fig:teaser}): it conflates direct and exhaustive
paths, and gives identical zeros to agents that never located evidence and those
stopping a single action short \citep{ma2024agentboard,michelakis2025core}.

Measuring these differences requires a defensible reference cost. Authored
solutions establish solvability but rarely prove minimality, while
cost-weighted evaluations require assumptions about the relative prices of
tools \citep{liu2025costbench}. DetourBench instead measures structural
interaction under uniform access costs. We restrict each task to a finite
transition system in which resources are partially observable but become
globally addressable after discovery: once an identifier is revealed, the
agent can invoke it directly without navigation or backtracking costs.
Exhaustive search over the complete transition graph then computes the minimum
number of charged accesses required for success, denoted $L_o$, and produces a
replayable shortest-path witness.

The same search supports failure-aware evaluation. For an unfinished episode
whose terminal state remains graph-recoverable, it computes the exact residual
work $R_T$: the minimum number of further accesses needed for success. This
allows success, oracle-relative attainment, normalized detour, and
residual-distance progress to be reported separately. An irreversible wrong
commitment is represented explicitly rather than assigned a finite residual
distance. $L_o$ is an offline optimum over the declared transition system, not
the cost of an optimal policy under partial observability; \S\ref{sec:method}
bounds what that leaves out.

DetourBench applies this measurement contract to two complementary
collections. The \emph{Controlled Mechanism Suite} holds graph structure,
targets, and oracle bounds fixed while varying a local information cue. The
\emph{Integrated Workflow Suite} combines multiple records, constraints, and
workflow actions in longer authored tasks. Both use the same transition
semantics, solver, evaluator, and artifact-versioning protocol. Mechanical
replay, provenance, necessity audits, pair diffs, and hash consistency
gate which artifacts and trajectories may be reported.

The measurement finds cases endpoint scoring cannot. On eight of twenty-five
tasks every model succeeds on every repeat, so success separates nothing at
all, and conditional attainment still spreads the models by 0.07 to 0.19 ---
three to nine times the variation across repeats of the same model. On the failing side, residual work
tells a trajectory that stopped one access short from one that exhausted its
budget far from the goal and from one that committed irreversibly --- three
situations a binary score records identically.

Our contributions are:
\begin{enumerate}\setlength{\itemsep}{0pt}\setlength{\parskip}{0pt}
  \item \textbf{Exact oracle-relative trajectory measurement.}
  We compute instance-specific full-information minima and replayable witnesses,
  together with exact residual work for graph-recoverable failures.
  \item \textbf{A shared benchmark and validation contract.}
  Controlled and integrated tasks use unified execution semantics and
  fail-closed checks for mechanical validity, provenance, pair integrity, and
  artifact consistency.
  \item \textbf{A diagnostic multi-model evaluation.}
  Across 375 episodes from five endpoints on 25 tasks, eight tasks are solved
  by every model on every repeat while cost still orders them, and residual
  work decomposes the failures a binary score records identically. \S\ref{sec:experiments} states
  the selection screen these tasks passed.
\end{enumerate}

DetourBench is deliberately formulated as a controlled diagnostic probe with exact graph-theoretic bounds, designed to isolate unnecessary and missing evidence accesses, rather than as a representative sample of natural workflows.

\section{Related Work}
\label{sec:related}

\paragraph{Agent and trajectory evaluation.}
Interactive-agent benchmarks commonly emphasize final task completion, while
process-oriented approaches inspect intermediate behavior. AgentBoard
\citep{ma2024agentboard} measures progress through annotated subgoals, and CORE
\citep{michelakis2025core} encodes each task as a deterministic automaton over
a manually verified set of acceptable tool-use paths, scoring correctness,
ordering, prefix criticality, harmful calls, and efficiency against it.
AgentRewardBench \citep{lu2025agentrewardbench} instead studies the reliability
of automatic trajectory evaluators. Both AgentBoard and CORE admit several
acceptable routes rather than one canonical ordering, so the difference is not
rigidity but provenance: their references are authored and establish that a
route is acceptable, while ours is search-derived and establishes that no
shorter one exists.

\paragraph{Interaction and path efficiency.}
Success weighted by Path Length (SPL) in embodied navigation
\citep{anderson2018spl} established the principle our attainment measure
inherits: compare an observed successful trajectory with an instance-specific
shortest path. Two things differ. SPL scores a failed episode zero, so it
cannot distinguish an agent that stopped one action short from one that never
located the evidence, and it has no representation for an irreversible wrong
commitment; those three cases are what our failure-side measurement separates,
using the same search that produces the bound. This is not a corner case on
our suite --- roughly a third of episodes fail on three of the five models,
and the ordering those failures induce differs from the ordering on success.
The shortest path is also over a different object: SPL's optimum is a distance
in a metric space, ours is over a combinatorial state of visited resources,
acquired credentials, and an outcome flag, in which prerequisites make
reachability history-dependent and one transition is absorbing.

CORE's efficiency measure is the closest contemporary comparison. It divides
the longest acceptable path not exceeding the agent's own length by that
length, which makes the reference adapt to the trajectory being scored and
leaves efficiency undefined for an agent shorter than every acceptable path.
Our reference is instead a fixed per-instance minimum that no legal route can
beat, so attainment is defined for every success and comparable across
episodes of very different length. CostBench
\citep{liu2025costbench} evaluates planning under heterogeneous tool prices and
dynamic failures, while work on efficient tool-integrated reasoning directly
optimizes models to reduce unnecessary calls \citep{wang2025actingless}.
DetourBench instead uses uniform access costs to isolate structural interaction
count; it does not measure latency, token consumption, or economic cost.

\paragraph{Partial observability and benchmark validity.}
Comparing observed trajectories to full-information optima connects to online
search and POMDPs \citep{papadimitriou1991,kaelbling1998planning}. Without
current-location costs, search depends on proximal cues or \emph{information
scent} \citep{pirolli1999}, motivating our controlled cue intervention. Audits
show underspecified tasks and scoring shortcuts threaten benchmark validity
\citep{abc2025,protocolvalidity2026,tian2026ignorance,swebenchverified}; DetourBench addresses
these via replayable witnesses, provenance, pair diffs, and cryptographic manifests.

\section{Methodology}
\label{sec:method}

\begin{figure*}[t]
\centering
\includegraphics[width=0.58\textwidth]{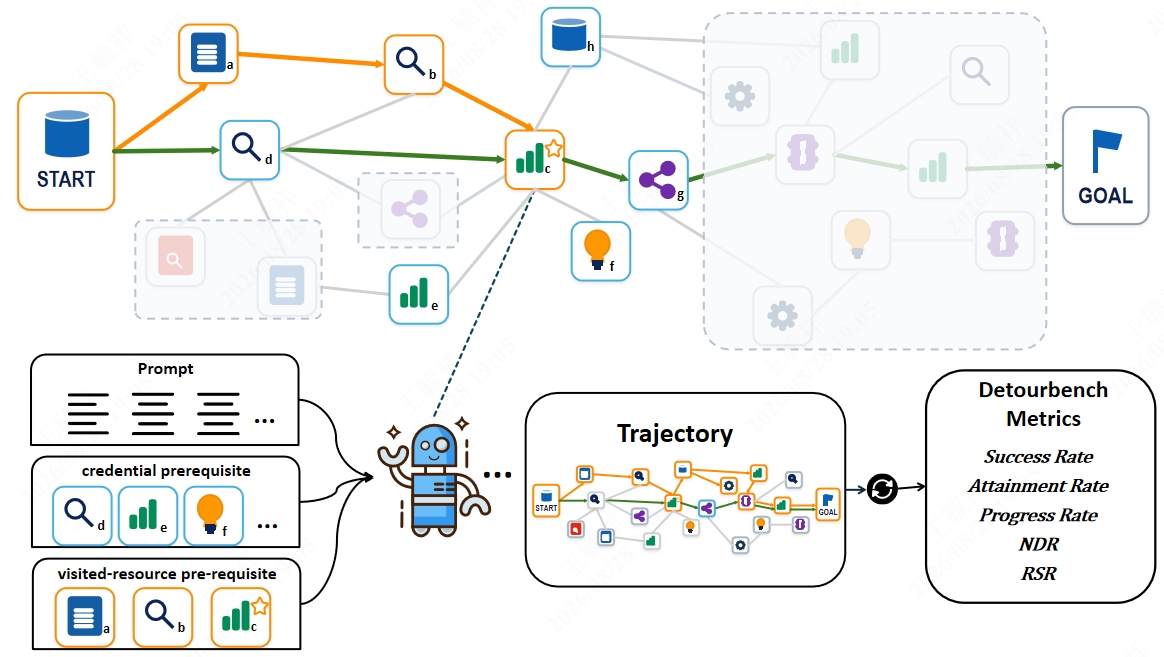}
\caption{\textbf{DetourBench transition semantics and evaluation pipeline.}
Agents interact with partially observable finite transition systems in which
discovered resource identifiers remain globally addressable. Exhaustive search
computes the full-information minimum charged accesses $L_o$ and residual work
$R_T$. Success, conditional attainment, residual progress, normalized detour,
redundant interaction, and irreversible failure are reported separately.}
\label{fig:framework}
\end{figure*}

\subsection{Measurement scope}

Figure~\ref{fig:framework} summarizes the semantics and the evaluation
pipeline. DetourBench measures structural interaction under uniform access
costs. For
each instance it compares an observed trajectory with the shortest successful
trajectory available to an evaluator that knows the complete transition system,
so the reference is a full-information offline optimum: exact under the
declared semantics, and not the cost of an optimal policy under partial
observability. Excess over it is therefore \emph{oracle-relative detour}, made precise
once the measures are defined.

\subsection{Transition and execution semantics}

An instance defines a finite action set, a deterministic transition function
$\delta$, and states
\begin{equation}
  s_t=(V_t,K_t,z_t),
  \label{eq:state}
\end{equation}
where $V_t$ is the set of accessed resources, $K_t$ is the set of acquired
credentials, and
$z_t\in\{\mathrm{active},\mathrm{success},\mathrm{wrong}\}$ is the outcome
state. Seed resources belong to $V_0$. Each environment action $a$ has a
source $\sigma(a)$, resource prerequisites $R_V(a)$, and credential
prerequisites $R_K(a)$. While the task is active, the enabled actions are
\begin{equation}
  \mathcal A(s_t)=
  \{a:\sigma(a)\in V_t,\;
       R_V(a)\subseteq V_t,\;
       R_K(a)\subseteq K_t\}.
  \label{eq:enabled-accesses}
\end{equation}

The agent observes only $o_t=\Omega(s_t)$. Accessing a resource may reveal
additional action identifiers or credentials. While $z_t=\mathrm{active}$,
both $V_t$ and $K_t$ grow monotonically, and previously enabled actions remain
available. Once an identifier has been revealed, it can be invoked directly;
the environment has no current-location or backtracking cost.

At model turn $u$, the agent may emit an ordered batch of candidate accesses.
The environment dispatches them sequentially until the batch ends, the access
budget is reached, or a terminal transition occurs. Let
$\tilde{\mathbf a}_u$ denote the prefix actually dispatched. The charged
trajectory and its cost are
\begin{equation}
  \tau=\tilde{\mathbf a}_1\oplus\tilde{\mathbf a}_2\oplus\cdots,
  \qquad
  C_\tau=\sum_u |\tilde{\mathbf a}_u|.
  \label{eq:trajectory-cost}
\end{equation}
Thus, cost counts dispatched environment accesses rather than model turns.
Every access reaching the environment dispatcher is charged, including
disabled, invalid, or duplicate accesses that leave the task state unchanged.
Pure model responses that dispatch no access are uncharged. A separate
predeclared model-turn limit prevents infinite no-access loops and does not
enter $C_\tau$.

Where a task contract offers a textual submission, it is uncharged and
contracted into the terminal state during search; in our suite, all completions
occurred through charged terminal actions. An episode ends at success, an absorbing
wrong commitment, the access budget, the model-turn limit, or an explicit final submission.

\subsection{Exact oracle and residual work}

Let $\mathcal G$ be the set of success states. Exhaustive breadth-first search
over the finite reachable state space computes
\begin{equation}
  L_o=\min_{\tau:s_0\rightarrow\mathcal G} C_\tau,
  \label{eq:oracle-length}
\end{equation}
where only charged accesses contribute to path cost. The search predecessor
map yields a shortest successful trajectory $\tau_o$, which is replayed under
the same transition function used for model episodes. The graph, success
conditions, oracle length, and replayable witness are frozen together. All
admitted instances satisfy $L_o\geq1$.

For an episode ending in an active state $s_T$, residual work is
\begin{equation}
  R_T=d^*(s_T,\mathcal G),
  \label{eq:residual-work}
\end{equation}
the minimum further charged accesses required for success. Resources and
credentials are monotone, so nothing already enabled becomes unavailable, and
the one-shot actions some instances declare are terminal: spending one leaves
the active state rather than stranding the episode inside it. Every active
terminal snapshot is therefore graph-recoverable. Consequently,
\begin{equation}
  0\leq R_T\leq L_o,
  \qquad
  C_\tau+R_T\geq L_o.
  \label{eq:residual-bounds}
\end{equation}
Successful episodes have $R_T=0$; no finite residual distance is assigned after
an absorbing wrong commitment. Graph-recoverability refers to the transition
system and does not imply sufficient remaining budget to execute the residual path.

\subsection{Trajectory measures}

Let $S_\tau\in\{0,1\}$ indicate terminal success. For successful episodes,
conditional attainment is
\begin{equation}
  A_\tau=
  \begin{cases}
    L_o/C_\tau, & S_\tau=1,\\
    \mathrm{undefined}, & S_\tau=0.
  \end{cases}
  \label{eq:attainment}
\end{equation}
Thus, $A_\tau\in(0,1]$ measures successful interaction relative to the
full-information minimum. It is reported jointly with success rather than
used as a stand-alone model score.

For successful or active graph-recoverable episodes, we define
\begin{equation}
  P_\tau=1-\frac{R_T}{L_o},
  \qquad
  \mathrm{NDR}_\tau=
  \frac{C_\tau+R_T-L_o}{L_o}.
  \label{eq:progress-ndr}
\end{equation}
Here $P_\tau\in[0,1]$ is normalized residual-distance reduction, while
$\mathrm{NDR}_\tau\geq0$ is excess charged interaction relative to the offline
optimum. For wrong-terminal episodes, $R_T$, $P_\tau$, and
$\mathrm{NDR}_\tau$ are undefined; their frequency is reported separately.

\paragraph{What an omniscient reference does and does not concede.}
$L_o$ is computed by a searcher that knows the whole graph, so it is fair to
ask whether NDR charges an agent for discovery no policy could avoid. The
construction rule in \S\ref{sec:construction} bounds how much of it can be:
evidence a decision requires is encoded as a prerequisite, so the solver
reaches the target only by traversing it. We verify this rather than assert
it: every annotated evidence node in the evaluated suite lies on the shortest
successful route, with no exceptions. What the oracle keeps is not
the ability to skip evidence but the ability to pick the right lookalike
candidate first. We do not net that advantage out, so NDR is excess relative
to hindsight-optimal selection rather than a claim that any particular access
was unreasonable; separating the two needs an informed reference policy the
released suite does not contain.

Finally, let $I_\tau$ be the number of dispatched accesses that are invalid
under the transition contract, and let $D_\tau$ be the number of valid
re-accesses that leave the whole state $s_t$ unchanged, outcome flag
included, so a state-advancing workflow action is never counted as a
repeat. The two sets are disjoint. For
successful episodes,
\begin{equation}
  \mathrm{RSR}_\tau=
  \frac{I_\tau+D_\tau}{C_\tau},
  \label{eq:rsr}
\end{equation}
with lower values indicating fewer invalid or duplicate accesses. RSR is
undefined on failure because early termination can produce a deceptively low
ratio. Unlike NDR, RSR captures only directly identifiable invalid or
state-preserving repetition.

\subsection{Access budgets}

The exact oracle is a reference cost rather than an episode limit. Each
instance also declares an information-complete witness of length $L_i$, which
exposes all resources marked necessary for an informed terminal decision. Its
frozen access budget is
\begin{equation}
  B=\left\lceil m\max(L_o,L_i)\right\rceil,
  \label{eq:budget}
\end{equation}
where the multiplier $m$ is fixed before the primary evaluation; we use $m=1.5$ throughout, so an instance with $L_o=12$ is run under a budget of 18. On every instance in the
released suite $L_i=L_o$, because necessity is enforced through prerequisites
that the shortest route must already satisfy; the maximum is therefore
inactive here and $B=\lceil mL_o\rceil$ in practice. We keep the general form
because an instance whose informed witness exceeds its oracle route would
otherwise receive a budget that forbids an informed decision. Neither $L_i$
nor $B$ is
used as an efficiency numerator. The release reports $L_o$, $L_i$, $m$, $B$,
and $B/L_o$ for every evaluated instance; no budget is increased during an
episode.

Of the 99 failures in the integrated suite, 95 exhaust this budget. They end a
median of three accesses from the goal (mean residual 3.02, mean progress
0.766), and 47 of them end within two accesses. Duplicate and invalid accesses
make up 7.6\% of their charged steps, so the budget is spent mostly on valid,
non-repeated accesses. This is consistent with resolving uncertainty among
lookalike candidates, but we do not separate reasonable uncertainty resolution
from avoidable exploration, and we do not vary $m$; budget sensitivity is left
to future work.

\section{Benchmark Construction and Validation}
\label{sec:construction}

\subsection{Collections}

A \emph{task} specifies a construction procedure, an \emph{instance} is one
frozen executable realization, and an \emph{episode} is one model run on an
instance. DetourBench contains two collections. The \emph{Controlled Mechanism
Suite} builds paired instances across six mechanisms; only the 40 Currency
pairs survived the equivalence audit in a form usable for a controlled
comparison (\S\ref{app:audits}), and they are what \S\ref{sec:experiments} reports.

The \emph{Integrated Workflow Suite} contains longer authored tasks grounded
in software, clinical, financial, banking, logistics, regulatory, legal,
corporate-registry, transit, customs, and security records. These tasks
combine multiple sources, scenario constraints, credentials, and workflow
actions. Each integrated task contributes one frozen instance, and the
twenty-five evaluated here are listed in Appendix~\ref{app:pertask}.

\subsection{Construction and controlled pairing}

Construction begins with source records rather than a prompt alone. A builder
emits the resource and action graph, prerequisites, terminal conditions, and
authored scenario facts. Source-derived content is linked to supporting spans,
while authored overlays are labeled separately. A removal audit deletes each
node on the oracle route and re-runs the exact search; for tasks 21--24 we also
execute content counterfactuals that change one fact and check that the
decision flips. Elsewhere semantic roles are documented, not executed
(Appendix~\ref{app:necessity}).

The solver computes $L_o$, generates a shortest-path witness, and replays it
under the model-facing evaluator. Each released task ships its graph, initial
state, task specification, source corpus, oracle, replayable witness, budget,
verifier report, and freeze manifest, and every trajectory records the graph
and oracle hashes it ran against.

A controlled pair holds source records, transition topology, actions, target,
answer, and oracle bound fixed while changing one declared cue sentence,
analogous to holding instructions fixed while toggling access to private task
conventions \citep{tian2026ignorance}. Canonical artifact diffs admit a pair only when the observed difference is
confined to that cue. Additional textual or structural differences cause the
pair to be excluded from controlled analysis.

\subsection{Validation and freeze}

Every reported artifact passes four gates: \textbf{mechanical validity}
requires schema checks and witness replay; \textbf{construct validity} requires
source provenance, labeled overlays, and a structural removal audit, with
documented semantic roles for most tasks and executed content counterfactuals
for four (Appendix~\ref{app:necessity});
\textbf{generation validity} checks uniqueness, answer degeneracy, and
controlled-pair integrity; and \textbf{artifact validity} freezes the graph,
oracle, prompt, metric, evaluator, and budget, with the graph and oracle
carrying cryptographic hashes that every trajectory records and every analysis
re-checks before reporting.

Development pilots identify malformed graphs, shortcut answers, ambiguous
prompts, and evaluator defects. A repair creates a new artifact version and
invalidates earlier trajectories. Because model behavior informed some
revisions, freezing guarantees artifact consistency but does not make the
benchmark an untouched test set. Experimental selection based on model
behavior is disclosed separately in Section~\ref{sec:experiments}. Analysis
fails closed on a version or hash mismatch, a runtime budget change, or an
incomplete model--task--repeat cell.

\section{Experiments and Results}
\label{sec:experiments}

We evaluate whether outcome-only scoring masks differences in execution
efficiency and partial progress. Two experiments do this from opposite
directions: a controlled intervention on the mechanism suite, where one
sentence is added or withheld and everything else is held fixed, and an
observational run over the integrated suite. They use different instances and are reported separately, never pooled;
qwen3.6-plus and glm-5.2 appear in both.

\subsection{A controlled cue intervention}

The mechanism suite lets one variable move. Both arms share graph, targets,
oracle bound, and correct answer, and the point-in-time view is an available
action with its URL on the edge in each; only the seed sentence naming it is
removed. We run 40 paired problems, four models, three repeats per arm: a balanced
$4\times40\times3\times2 = 960$-episode design, 120 episodes per model per
arm; Table~\ref{tab:signpost} reports both arms.

\begin{table}[t]
\centering\small
\setlength{\tabcolsep}{4pt}
\caption{Controlled intervention. Query is the fraction of episodes that enter
the point-in-time view. Waste is the mean number of out-of-scope provisions
read. Deleting one sentence is the only difference between the arms.}
\label{tab:signpost}
\begin{tabular}{lrrrrcrrrr}
\toprule
& \multicolumn{4}{c}{signposted} && \multicolumn{4}{c}{unsignposted} \\
\cmidrule(lr){2-5}\cmidrule(lr){7-10}
Model & Query & Pass@1 & Pass$^3$ & Waste && Query & Pass@1 & Pass$^3$ & Waste \\
\midrule
qwen3.6-plus     & 100\% & 98\% & 95\% & 0.0 &&  5\% & 42\% & 32\% & 4.2 \\
deepseek-v3.2    &  96\% & 69\% & 40\% & 1.4 && 68\% & 38\% &  8\% & 3.2 \\
glm-5.2          &  87\% & 71\% & 48\% & 1.9 && 32\% & 37\% & 32\% & 4.3 \\
claude-opus-4-8  &  77\% & 85\% & 62\% & 1.0 &&  2\% & 42\% & 40\% & 4.2 \\
\bottomrule
\end{tabular}
\end{table}

Across 160 pairs, removing the sentence drops view use by 63 points and
success by 41, while raising out-of-scope reads by 2.9 per episode. In view
use, 131 of the 160 model--problem pairs move toward the unsignposted arm, 27
do not move, and 2 move the other way.
Successful episodes take longer routes: conditional attainment falls from 0.826
to 0.644 on Qwen and 0.794 to 0.629 on Claude. The endpoint was never withheld,
only the sentence pointing at it.

$\mathrm{pass}^k$ separates two failure modes $\mathrm{pass}@1$ conflates:
in the unsignposted arm deepseek-v3.2 holds 38\% under $\mathrm{pass}@1$ but
8\% under $\mathrm{pass}^3$, landing the answer occasionally rather than
reliably, while claude-opus-4-8 falls less under $\mathrm{pass}^3$, so its
failures are a stable per-problem property.

In the signposted arm 32 of the 40 problems pass above 70\% against 13 in
the unsignposted arm, so the signposted arm is near a ceiling for current
models and the intervention stays measurable because the other one is not.

Because the intervention changes which episodes succeed, we also compare
attainment only where both arms succeed: on the $n=176$ model--problem--repeat
pairs that succeed in both arms, drawn from 31 problems, withholding the
signpost lowers conditional attainment from 0.763 to 0.637 ($\Delta = -0.126$;
95\% problem-cluster bootstrap interval $[-0.156, -0.099]$, 20{,}000 resamples
of problems with all their pairs, seed 20261107) and raises
out-of-scope reads from 0.91 to 2.72. Averaged per problem, the decline holds
on 23 of the 31 problems ($t_{30}=-6.1$).

\subsection{Experimental setup}

The Integrated Workflow Suite contains 25 versioned tasks: 20 frozen before
review (tasks 1--20), four authored afterwards (tasks 21--24), and one run
alongside tasks 1--20 but left out of their manifest (task 25). We evaluate five
model endpoints (\texttt{claude-opus-5}, \texttt{deepseek-v4-pro},
\texttt{qwen3.6-plus}, \texttt{glm-5.2}, \texttt{kimi-k3}) attempting every
task three times through a shared scaffold ($25\times5\times3=375$ episodes).

Fixed budgets are enforced without runtime unlocking. Infrastructure aborts
were rerun; all valid, invalid, and duplicate accesses are charged. In this
suite, all completions occurred through charged terminal accesses. All 375 runs
carry cryptographic graph and oracle hashes; six tasks (17, 20, and 21--24) ran
under a later prompt revision (Appendix~\ref{app:pertask}). The task is the unit
of comparison: repeats characterize within-task variability, not 75 independent
samples.

\subsection{Outcome and successful-run efficiency}

Table~\ref{tab:e5models} reports endpoint success together with attainment
and RSR on successful episodes. These conditional metrics should be read
jointly with success because models need not succeed on the same tasks.

\begin{table}[t]
\centering
\small
\setlength{\tabcolsep}{4pt}
\caption{Results on the 25-task Integrated Workflow Suite ($N=375$). Conditional
attainment and RSR are averaged over successful episodes (best in bold; lower RSR is better). Model identifiers correspond to endpoints in Section~\ref{sec:experiments}.}
\label{tab:e5models}
\begin{tabular}{lrrrrr}
\toprule
Model & Episodes & Successes & Success rate &
Cond.\ attainment & Succ.\ RSR \\
\midrule
Claude
  & 75 & \textbf{72} & \textbf{96.0\%} & \textbf{0.890} & 0.026 \\
Kimi
  & 75 & 69 & 92.0\% & 0.861 & \textbf{0.023} \\
DeepSeek
  & 75 & 46 & 61.3\% & 0.803 & \textbf{0.023} \\
Qwen
  & 75 & 45 & 60.0\% & 0.806 & 0.040 \\
GLM
  & 75 & 44 & 58.7\% & 0.836 & 0.034 \\
\bottomrule
\end{tabular}
\end{table}

Across models, 276 of 375 episodes succeed. Claude and Kimi obtain the
highest success rates and conditional attainment. Among the remaining
models, however, outcome and efficiency orderings differ: GLM has the
lowest success rate but higher conditional attainment than DeepSeek and
Qwen. This is a descriptive ranking reversal because conditional attainment is
computed on model-specific successful subsets. Averaging per task rather than per
episode keeps GLM above both (0.811 vs.\ 0.793 for Qwen and 0.789 for DeepSeek).
Restricting to the thirteen tasks solved by all models again puts GLM third (0.851
vs.\ 0.829 for Qwen and 0.824 for DeepSeek). Qwen and DeepSeek differ by at most
0.005 and swap on the nineteen tasks run under the original prompt, so we do not
order them.

Successful-run RSR ranges from 0.023 to 0.040. Its ordering differs from
both success and attainment because RSR isolates invalid and duplicate
accesses, whereas attainment captures all charged work beyond the Exact
oracle.

\subsection{Matched success with different costs}

Eight tasks---2, 3, 4, 5, 6, 14, 19, and 25---are solved in all three
repeats by all five models. Their between-model attainment spreads range
from 0.074 to 0.190. Because every episode succeeds, these differences
cannot be attributed to unequal success counts.

The widest is task 3 ($L_o=15$, spread 0.190), where mean accesses run from 17.0
to 21.7 across models. The extra cost is not repetition: the best and worst
models both record $\text{RSR}=0.000$ (zero repeats or invalid calls) yet differ
in attainment by 0.189. RSR detects overt repeats; attainment also catches valid
but unnecessary exploration (Appendix~\ref{app:task3}). These tasks establish
cost variation exists, not prevalence, as separation was a screening criterion.

\subsection{Failure-aware analysis}

Of the 375 scored episodes, 99 fail: 95 exhaust the access budget, two stop
in a recoverable state, and two enter an absorbing wrong state.
Table~\ref{tab:failure-analysis} (Appendix~\ref{app:failure}) reports progress and NDR over the 97
failures that remain active and recoverable. Wrong-terminal episodes are
excluded because they have no finite residual distance.

These columns are not a model comparison: the five models fail on largely
disjoint task sets (Claude on 1 instance, Kimi on 4, DeepSeek on 13, GLM on 14,
Qwen on 16). Across all 97 recoverable failures, progress ranges from well short
of the goal to one access away, and the two absorbing commitments have no
residual distance at all. On task 18 ($L_o=20$), six failures collapse to one
binary score; our decomposition separates two that stopped one access from
success, three that exhausted budget at 0.85--0.95 progress, and one absorbing
commitment with no residual distance (Appendix~\ref{app:task18}).

\subsection{Selection boundary}

The 20-task manifest was frozen on August 28, 2026; model behavior informed
authoring for eight tasks and replacement of two uninformative instances. Tasks
21--24 were authored after review; two of them (21 and 22) were revised after
earlier model runs, and only the final revision, on which all five models were
run, is reported. Task 25 was run with tasks 1--20 under the original prompt and
passes the same screen, but was left out of the frozen manifest; we add it here.
The 8/17 cost/outcome split reflects selection choices rather
than prevalence.
Retention rules, excluded candidates, and hashes are in Appendix~\ref{app:audits}.

\section{Conclusion and Limitations}
\label{sec:conclusion}

DetourBench evaluates interactive agents through trajectory-level transition systems
with provable minimum-action bounds $L_o$ and exact residual work $R_T$, preventing
efficient success, exploratory success, recoverable near-misses, and irreversible
error from collapsing into one score. Experiments confirm this distinction is
practical: one withheld sentence moves the same model on the same task, eight tasks
separate models purely by cost under 100\% success, and binary failure scores
decompose into near-complete, costly-incomplete, and irreversible cases.

Four limitations bound this: model-informed screening reflects released suite
properties rather than population prevalence; three repeats limit stochastic
analysis; the omniscient oracle does not net out lookalike selection
advantage; and exactness holds under authored semantics, whose semantic
necessity is executed for only four tasks. Future work should
evaluate pre-registered held-out suites, widen repeat and model coverage, add
human baselines, and develop informed reference policies separating discovery
from planning error.

{\small
\setlength{\bibsep}{0pt}
\bibliographystyle{plainnat}
\bibliography{references}
}

\newpage
\appendix
\section{Complete Integrated Task Graph}
\label{app:example-graph}

\begin{figure}[H]
  \centering
  \includegraphics[width=\textwidth]{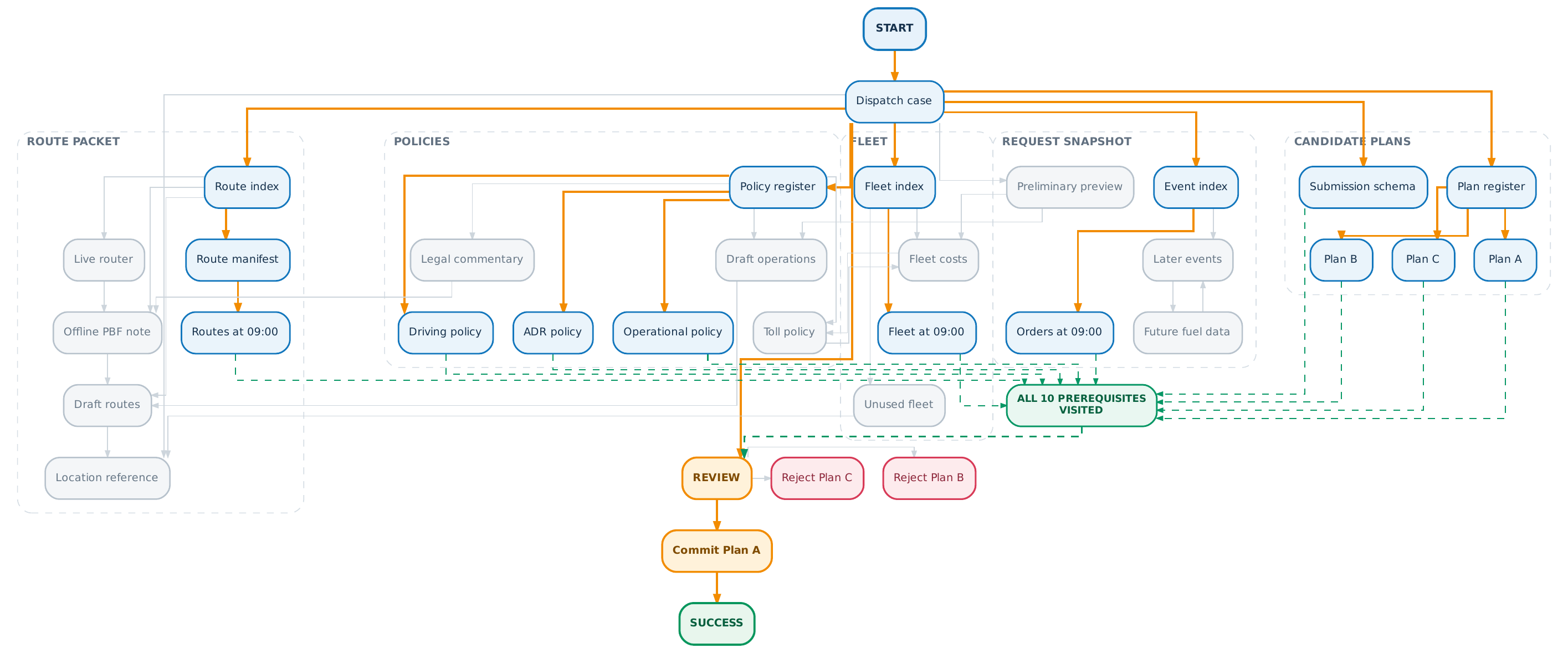}
  \caption{\textbf{Complete graph for Task 18 (freight-dispatch cutoff, $L_o=20$).}
  All 35 nodes and 45 discovery edges are shown. Blue nodes and orange edges
  are accessed by one replayed shortest witness ($L_o=20$); pale gray nodes
  and edges are reachable optional or distractor resources. Green dashed
  connectors unpack the review action's ten visited-resource prerequisites.
  Committing Plan A reaches success, whereas committing Plan B or Plan C is an
  absorbing wrong transition. Because discovered identifiers remain globally
  addressable, the witness may alternate among evidence families without
  incurring a navigation or backtracking cost. The rendering is generated
  directly from the released graph and oracle; labels are shortened, but no
  nodes or discovery edges are omitted.}
  \label{fig:freight-graph}
\end{figure}

\section{Construction and selection audits}
\label{app:audits}

Construction audits materially restricted which artifacts were eligible for
analysis. The counts below are paired artifacts as they stood at audit time,
before repair and freeze, and each pair is two renderings of one problem.
Table~\ref{tab:audit-summary} summarizes the principal findings and the
corresponding corrective actions. Rejected artifacts remain available in
the release but are not used to support the reported comparisons.

\begin{table}[H]
\centering
\small
\caption{Summary of benchmark-construction audits.}
\label{tab:audit-summary}
\begin{tabular}{p{0.18\linewidth}p{0.43\linewidth}p{0.29\linewidth}}
\toprule
Audit & Finding & Action \\
\midrule
Paired-arm
equivalence
&
All 53 Currency pairs and 30 of 32 Breadth pairs passed content
equivalence; the controlled experiment reports the 40 Currency pairs every
model was run on. All 142 Sufficiency and Disposition pairs failed because
their arms differed beyond the intended intervention.
&
Only equivalence-validated pairs remain eligible for controlled
comparisons.
\\[3pt]

Instance
diversity
&
Eight nominal seeds initially produced only five distinct underlying
problems across 40 artifacts.
&
Case-specific variables were incorporated into generation and distinctness
was re-audited.
\\[3pt]

Answer
balance
&
Early Sufficiency and Disposition generators produced a fixed correct
answer direction.
&
Both answer directions are now instantiated and checked during validation.
\\[3pt]

Scoring
normalization
&
Exact-string matching treated equivalent citation formats as different
answers, producing 38 false negatives among 61 inspected episodes.
&
Structured outputs are canonicalized before semantic comparison.
\\
\bottomrule
\end{tabular}
\end{table}

The formal 20-task evaluation set was selected after observing model
behavior. Two instances that separated the models by neither measure were replaced by tasks 1 and 14. A task was
retained if models differed in empirical success rate or, when all models
succeeded, if the between-model conditional-attainment spread was at least
twice the mean within-model standard deviation across the three repeats.
This rule and the excluded candidates are recorded in
\texttt{paper\_selection\_20.json}. Tasks 21--24 were authored after review
and did not pass through this screen; each separates the models by outcome.
Task 25 was evaluated in the same August runs as tasks 1--20 but is not in
that manifest; it passes the screen by cost (attainment spread 0.074 against a
mean within-model standard deviation of 0.020, ratio 3.8). Tasks 21--25 are
released with their own frozen graphs, oracles, and hashes.

\section{Necessity audits: what is checked}
\label{app:necessity}

\paragraph{Structural removal.} For every task we delete each node on the
oracle route (all edges into it and out of it) and re-run the exact search of
Appendix~\ref{app:solver_complexity}. Of the 314 such nodes across the 25
tasks, removing 309 leaves the target unreachable. The other five have
substitutes: one section in task 1 (the bound rises from 9 to 10), three records
in task 9 (bounds 13, 11, and 11), and one search page in task 19 (bound
unchanged at 16). This check shows that the route needs these nodes; it does not
show that their content matters to the decision, because necessity is enforced
through prerequisites.

\paragraph{Executed content counterfactuals.} Tasks 21--24 each ship a
frozen \texttt{necessity\_audit.json} in which a domain decision function is
re-run on edited inputs. Each task runs four edits: change one key fact, change
a relation or scope, remove a required piece of evidence, and, as a negative
control, replace the distractor records. In all four tasks every edit gives the
expected result, and an independent reference implementation agrees on the
baseline and all four edits. In task 23 (customs ruling revocation), for
example, the baseline determination is reliance protection at a 5.3\% duty
rate. Moving the entry date from 1993-04-15 to 1993-06-01, or setting good-faith
reliance to false, changes it to the revocation tariff at 20.0\%. Removing
required evidence leaves no determination. Replacing the twelve distractor
records leaves the baseline unchanged. These counterfactuals are authored edits
to the inputs, not revised observations, and they cover a finite set of edits
rather than every node.

\paragraph{Documented roles.} For tasks 8--20 the release records an authored
semantic role for each evidence node or edge, without an executed content
counterfactual; tasks 1--7 and 25 carry the structural check only. For these
21 tasks semantic necessity rests on the authored scenario, which is part of
the limitation stated in Section~\ref{sec:conclusion}.

\section{Task 3, transitive policy check: per-model costs}
\label{app:task3}

All five models solve task 3 on every repeat, so endpoint success assigns
them the same score. Their charged accesses do not agree, and neither does what the extra
accesses consist of (Table~\ref{tab:task3}).

\begin{table}[t]
\centering\small
\caption{Task 3, exact oracle $L_o=15$, three repeats per model. RSR counts
invalid and duplicate accesses only.}
\label{tab:task3}
\begin{tabular}{lrrr}
\toprule
Model & Mean charged & Attainment & RSR \\
\midrule
Claude   & 17.0 & \textbf{0.882} & \textbf{0.000} \\
Kimi     & 17.7 & 0.850 & 0.019 \\
Qwen     & 21.0 & 0.717 & 0.014 \\
DeepSeek & 21.3 & 0.705 & 0.015 \\
GLM      & 21.7 & \textbf{0.693} & \textbf{0.000} \\
\bottomrule
\end{tabular}
\end{table}

The two models with an RSR of exactly zero are the most and least efficient on
this task. Neither repeats an access or issues an invalid one, so a redundancy
measure rates them identically, while their attainment differs by 0.189: GLM
spends 6.7 accesses beyond the bound on unrelated dependencies, rejected
remediation candidates, and release metadata, all of them valid and none of
them repeated. That gap is invisible to RSR and is exactly what an
oracle-relative measure is for.

\section{Failure decomposition across models}
\label{app:failure}

Table~\ref{tab:failure-analysis} gives the per-model counts
behind the decomposition reported in \S\ref{sec:experiments}.

\begin{table}[t]
\centering
\small
\setlength{\tabcolsep}{4pt}
\caption{Failure decomposition. Progress and NDR are means over recoverable
failures only, and \emph{Tasks} is how many distinct instances those failures
come from. The columns are not a model comparison: the five sets of failing
instances are largely disjoint and no instance yields a recoverable failure
from all five models, so nothing here is ranked.}
\label{tab:failure-analysis}
\begin{tabular}{lrrrrrr}
\toprule
Model & Failures & Recoverable & Tasks & Wrong & Progress & NDR \\
\midrule
Claude   &  3 &  3 &  1 & 0 & 0.519 & 1.037 \\
Kimi     &  6 &  6 &  4 & 0 & 0.830 & 0.678 \\
DeepSeek & 29 & 29 & 13 & 0 & 0.712 & 0.791 \\
Qwen     & 30 & 28 & 16 & 2 & 0.779 & 0.727 \\
GLM      & 31 & 31 & 14 & 0 & 0.829 & 0.678 \\
\midrule
Total    & 99 & 97 & -- & 2 & 0.770 & 0.737 \\
\bottomrule
\end{tabular}
\end{table}

\section{Task 18 failure decomposition}
\label{app:task18}

Task 18 makes the distinction concrete. Its exact oracle requires
$L_o=20$ accesses, with a budget of 30. Claude and Kimi succeed in all
three repeats, using 21.7 and 23.7 accesses on average. DeepSeek, Qwen, and
GLM each succeed once, but their failed trajectories terminate in different
states.

DeepSeek and GLM each produce one recoverable \texttt{gave\_up} trajectory
after reaching \texttt{committed\_a}, only one required access from success.
Both have progress 0.95, with NDR values of 0.30 and 0.35, respectively.
Their budget-exhausted runs are less efficient: DeepSeek ends with progress
0.85 and NDR 0.65, while GLM reaches progress 0.95 but incurs NDR 0.55
through additional exploration. Qwen likewise has one budget-exhausted run
with progress 0.85 and NDR 0.65, but its other failure commits to
\texttt{rejected\_b}, an absorbing wrong state for which residual metrics
are undefined.

Binary scoring assigns all six trajectories the same failure value. The
graph-based decomposition instead separates a one-step-away recoverable
execution, inefficient budget exhaustion, and an irreversible incorrect
commitment.

\section{Per-task evaluation results}
\label{app:pertask}

Table~\ref{tab:e5tasks} reports the 25 evaluation tasks, 1--20 in manifest
order followed by the four authored after review and task 25.
Success spread is the difference between the largest and smallest model
success rates. Attainment spread is computed across model-level conditional
attainment values that are defined for the task. The table reports raw
descriptive spreads rather than normalizing them by a standard deviation
estimated from only three repeats.

\begin{table}[t]
\centering
\small
\setlength{\tabcolsep}{4pt}
\caption{Per-task results. Success is aggregated over five models and three
repeats. $\dagger$ marks the six tasks evaluated under the later prompt
revision; dropping them and recomputing on the remaining nineteen leaves the
success ordering unchanged and the cost ordering unchanged except that Qwen and
DeepSeek, which we do not order, swap. Tasks 21--24 were added after review;
task 25 was run with tasks 1--20 but omitted from their manifest. For tasks 1--20 the released manifest
\texttt{paper\_selection\_20.json} maps each ID to its artifact directory,
graph and oracle hashes, task revision, and prompt version; it also records each task's pre-renumbering build number, which the
artifact directory names carry. Hashes remain the authoritative identity.}
\label{tab:e5tasks}
\begin{tabular}{rlllrrrr}
\toprule
ID & Task & Domain & Separates by & $L_o$ & Success & succ$\Delta$ & att$\Delta$ \\
\midrule
 1 & Fall-protection standard    & Regulatory & outcome &  9 &  7/15 & 1.00 & 0.107 \\
 2 & Direct advisory fix         & Software   & cost    & 10 & 15/15 & 0.00 & 0.091 \\
 3 & Transitive policy check     & Software   & cost    & 15 & 15/15 & 0.00 & 0.190 \\
 4 & Multi-parent remediation    & Software   & cost    & 15 & 15/15 & 0.00 & 0.118 \\
 5 & Multiple safe versions      & Software   & cost    & 13 & 15/15 & 0.00 & 0.147 \\
 6 & Conflicting fix evidence    & Software   & cost    & 15 & 15/15 & 0.00 & 0.130 \\
 7 & Thiazide refill provenance  & Clinical   & outcome & 12 &  6/15 & 1.00 & 0.028 \\
 8 & Warfarin INR conflict       & Clinical   & outcome & 11 &  8/15 & 1.00 & 0.250 \\
 9 & Opioid refill packet        & Clinical   & outcome & 11 &  8/15 & 1.00 & 0.249 \\
10 & NSAID class review          & Clinical   & outcome & 12 & 12/15 & 0.67 & 0.225 \\
11 & ACE inhibitor in pregnancy  & Clinical   & outcome & 12 & 14/15 & 0.33 & 0.205 \\
12 & Encounter episode boundary  & Clinical   & outcome & 12 &  7/15 & 1.00 & 0.150 \\
13 & Restatement reliance        & Finance    & outcome & 12 & 11/15 & 1.00 & 0.113 \\
14 & Reissued card lifecycle     & Banking    & cost    & 17 & 15/15 & 0.00 & 0.139 \\
15 & Card PIN reset              & Banking    & outcome & 14 & 13/15 & 0.67 & 0.188 \\
16 & Settled comment letter      & Finance    & outcome & 12 & 11/15 & 0.67 & 0.064 \\
17 & Pinned upgrade bundle$^\dagger$ & Software   & outcome & 10 &  7/15 & 1.00 & 0.079 \\
18 & Freight dispatch cutoff     & Logistics  & outcome & 20 &  9/15 & 0.67 & 0.209 \\
19 & Periodontal onset date      & Clinical   & cost    & 16 & 15/15 & 0.00 & 0.109 \\
20 & Boundary dispute triage$^\dagger$ & Legal      & outcome & 19 & 10/15 & 1.00 & 0.259 \\
\midrule
21 & Lapsed-entity consolidation$^\dagger$ & Corporate & outcome & 14 &  9/15 & 1.00 & 0.075 \\
22 & Weekend timetable exception$^\dagger$ & Transit   & outcome & 14 & 11/15 & 1.00 & 0.122 \\
23 & Ruling revocation reliance$^\dagger$  & Customs   & outcome & 14 &  7/15 & 1.00 & 0.263 \\
24 & Build-tag CVE exemption$^\dagger$     & Security  & outcome & 14 &  6/15 & 1.00 & 0.019 \\
25 & PyPI multi-parent remediation & Software  & cost    & 16 & 15/15 & 0.00 & 0.074 \\
\bottomrule
\end{tabular}
\end{table}

The eight tasks labeled \emph{cost} have zero success spread because every
model succeeds in every repeat. The remaining seventeen tasks separate models
by endpoint outcome. Since this categorization follows the model-informed
screen described above, the resulting 8/17 split is a property of the
released evaluation set rather than a population estimate.

The release contains the frozen graphs, oracles, and artifact hashes, and the
harness that reruns every episode against a configured endpoint; recorded
model trajectories are not part of the public release. The harness sets no
temperature or top-$p$, so endpoints sample at provider defaults, and it stops
an episode after 60 model turns; no reported episode reached that limit (the
longest used 24). Tasks 1--20 and 25 were run on August 27--29, 2026, and
tasks 21--24 on November 4--5, 2026.

\section{Additional endpoint: HY4-Preview}
\label{app:hy_evaluation}

After the main runs were frozen we ran one further endpoint,
\texttt{HY4-Preview}, on tasks 1--24 with three repeats (72 episodes; it was
not run on task 25), against
the same frozen graphs, oracles, bounds, and budgets; every episode's graph and
oracle hashes match the released artifacts. All 72 episodes used the later
prompt revision marked $\dagger$ in Table~\ref{tab:e5tasks}. That revision
rewords the system prompt and also the not-ready message, which now tells the
agent that prerequisite records remain unread. The comparison with
Table~\ref{tab:e5models} is therefore not like-for-like on the 18 unmarked
tasks, and we report HY4-Preview separately rather than ranking it.

\begin{table}[H]
\centering\small
\caption{HY4-Preview on tasks 1--24 (3 repeats, later prompt revision).
Columns as in Table~\ref{tab:e5models}; progress and NDR are over recoverable
failures, as in Table~\ref{tab:failure-analysis}.}
\label{tab:hy_comparison}
\begin{tabular}{lrrrrrrr}
\toprule
Model & Episodes & Successes & Success rate & Cond.\ att. & Succ.\ RSR & Progress & NDR \\
\midrule
HY4-Preview & 72 & 52 & 72.2\% & 0.821 & 0.025 & 0.819 & 0.696 \\
\bottomrule
\end{tabular}
\end{table}

All 20 failures exhaust the budget; none submits a wrong determination, so all
are recoverable, and they fall on 12 of the 24 tasks. On the seven of these
tasks that every main-run model solves on every repeat (task 25 is the
eighth), HY4-Preview succeeds in 20 of 21 episodes.

\section{Cost of the exact oracle}
\label{app:solver_complexity}

We timed the exact search on the 25 frozen task graphs in single-threaded
CPython 3.10 on an Intel Xeon server. Runtime is the median of five runs; peak
memory is measured with \texttt{tracemalloc} in one further run; states are the
distinct (visited set, inventory) states the breadth-first search generates
before it reaches the target. All 25 recomputed bounds equal the frozen $L_o$.

\begin{table}[H]
\centering\small
\caption{Cost of the exact oracle across the 25 frozen tasks.}
\label{tab:solver_cost}
\begin{tabular}{lrrrr}
\toprule
Quantity & Minimum & Median & Mean & Maximum \\
\midrule
Graph nodes             & 16   & 30    & 33.7  & 62 \\
Search states generated & 12   & 87    & 1{,}272 & 13{,}541 \\
Runtime (ms)            & 0.07 & 0.85  & 20.3  & 273.4 \\
Peak memory (MiB)       & 0.02 & 0.09  & 1.74  & 23.97 \\
\bottomrule
\end{tabular}
\end{table}

The most expensive instance is task 20 (13{,}541 states, 273\,ms, 24\,MiB).
These figures describe the released graphs; because the search state is a set
of visited resources, they are not a bound for substantially larger graphs.

\newpage

\end{document}